\UseRawInputEncoding
\documentclass[sigconf, nonacm]{acmart}
\AtBeginDocument{%
  }


\definecolor{vova}{rgb}{0.1, 0.6, 0.2}
\usepackage{tcolorbox}
\usepackage{listings}
\usepackage{paralist}
\usepackage{makecell}

\newcommand{\method}{LLM4SFC}

\tcbuselibrary{listings,skins,breakable}

\tcbuselibrary{listingsutf8}           
\begin{document}

\title{\method: Sequential Function Chart Generation via Large Language Models}

\author{Ofek Glick}
\orcid{0009-0002-3458-4473}
\affiliation{
   \institution{Bosch Research}
   \city{Haifa}
   \country{Israel}
}
\email{ofek.gluck@gmail.com}

\author{Vladimir Tchuiev}
\orcid{0000-0002-5935-8168}
\affiliation{
  \institution{Bosch Research}
  \city{Haifa}
  \country{Israel}}
\email{vladimir.tchuiev@bosch.com}

\author{Marah Ghoummaid}
\orcid{0009-0001-7096-8944}
\affiliation{
  \institution{Bosch Research}
  \city{Haifa}
  \country{Israel}}
\email{marahghoummaid@gmail.com}

\author{Michal Moshkovitz}
\orcid{0009-0006-9456-9944}
\affiliation{
  \institution{Bosch Research}
  \city{Haifa}
  \country{Israel}}
\email{michal.moshkovitz@bosch.com}

\author{Dotan Di-Castro}
\orcid{0009-0001-0900-3932}
\affiliation{
  \institution{Bosch Research}
  \city{Haifa}
  \country{Israel}
}
\email{dotan.dicastro@bosch.com}

\begin{abstract}
While Large Language Models (LLMs) are increasingly used for synthesizing textual PLC programming languages like Structured Text (ST) code, other IEC 61131-3 standard graphical languages like Sequential Function Charts (SFCs) remain underexplored. Generating SFCs is challenging due to graphical nature and ST actions embedded within, which are not directly compatible with standard generation techniques, often leading to non-executable code that is incompatible with industrial tool-chains

In this work, we introduce \textsc{\method}, the first framework to receive natural-language descriptions of industrial workflows and provide executable SFCs. \textsc{\method} is based on three components: \textbf{(i)} A \emph{reduced structured representation} that captures essential topology and in-line ST and reduced textual verbosity; \textbf{(ii)} \emph{Fine-tuning and few-shot retrieval-augmented generation (RAG)} for alignment with SFC programming conventions; and \textbf{(iii)} A \emph{structured generation} approach that prunes illegal tokens in real-time to ensure compliance with the textual format of SFCs.

We evaluate \textsc{\method} on a dataset of real-world SFCs from automated manufacturing projects, using both open-source and proprietary LLMs. The results show that \textsc{\method} reliably generates syntactically valid SFC programs effectively bridging graphical and textual PLC languages, achieving a generation generation success of 75\% - 94\%, paving the way for automated industrial programming.

\end{abstract}


\maketitle

\section{Introduction}
\begin{figure}
    \centering
    \includegraphics[width=1\linewidth]{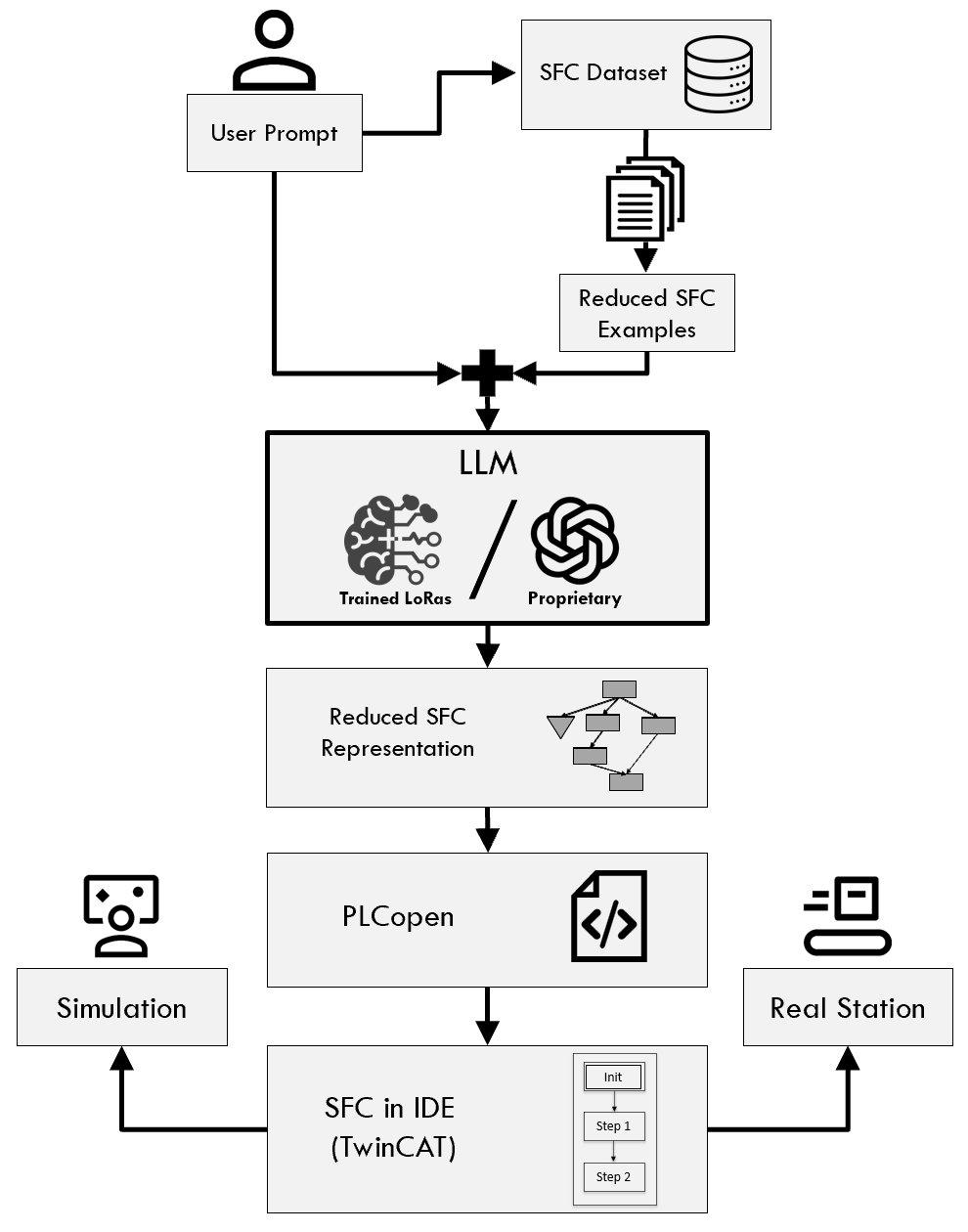}
    \caption{Overview of the \method generation pipeline. The user provides a natural language prompt describing the desired scenario. Semantically similar SFCs are embedded into the model's context as few-shot demonstrations. The LLM then generates a reduced representation of the SFC. This reduced representation encodes the structural topology of the control logic while abstracting away redundant metadata. The reduced SFC is parsed into the PLCopen XML format. The final output is a vendor compatible SFC diagram that can be viewed or edited in standard IDEs such as TwinCAT.}
    \Description[Generation pipeline overview]{Overview of the \method generation pipeline. The user provides a natural language prompt describing the desired scenario. Semantically similar SFCs are embedded into the model's context as few-shot demonstrations. The LLM then generates a reduced representation of the SFC. This reduced representation encodes the structural topology of the control logic while abstracting away redundant metadata. The reduced SFC is parsed into the PLCopen XML format. The final output is a vendor compatible SFC diagram that can be viewed or edited in standard IDEs such as TwinCAT.}
    \label{fig:LLM4SFC-pipeline}
    \Description{A diagram showing the overview of the LLM4SFC pipeline.}
\end{figure}

Programmable Logic Controllers (PLCs) are central to industrial automation, controlling critical processes across sectors such as such as process automation, manufacturing, energy, and transportation. The global PLC market is forecasted to surpass USD 15 billion by 2029 \cite{plc-forcast}, showcasing their widespread use. Despite their essential role, PLC programming remains highly labor intensive, particularly when using mixed language programming. Field surveys indicate that the software programming can consume up to 50\% of total system development efforts \cite{success-in-automation}.  Given this substantial resource demand, automation of PLC programming, particularly through code generation, is a crucial strategy for reducing development time and improving efficiency.

Recent works in code generation via large language models (LLMs) \cite{qwen2.5-coder, openai_chatgpt_2025, google_gemini_2025}, have demonstrated promising capabilities in generating complex functional programs in mainstream programming languages. In the context of PLC programming, prior research has largely focused on generating Structured Text (ST), leveraging the syntactic similarity of ST to high-level languages like Pascal and C \cite{agent4plc, LLM4PLC, LLMS-RL4PLC}. While these efforts have shown promising results, they often omit support for graphical languages, such as \textbf{\textit{Sequential Function Charts (SFCs)}}, which are still extensively used in industrial environments. 

While SFCs are typically visual, PLC development environments supports a structured textual representation of SFCs \cite{plcopen-xml}. However, this representation is not meant to be human readable and serves as an industry standard to make SFCs compatible between different IDEs.
The main challenge of generating SFCs lies in two factors. First, the textual representation of a graphical SFC is verbose, it embeds large amounts of metadata needed for compilability and machine execution, obscuring the semantics a model actually needs to learn. On top of that, other textual programming languages such as ST, which by itself is challenging to generate zero-shot \cite{LLM4PLC}, could be embedded in the SFC, only adding difficulty. Second, publicly available training corpora contain very few SFC examples because most industrial charts remain proprietary. As a result, large language models currently performed less effectively on SFC generation \cite{ascii-sfcs} compared to more widely used programming languages like Python or Java.

Treating SFCs as textual programmings rather then graphical, alongside improvements in the LLMs themselves are advancing SFC generation. First, schema-guided decoding enable models to emit strongly structured responses that map cleanly onto the rigid data formats expected by industrial tool-chains. Second, context windows now extend to hundreds of thousands of tokens, permitting a single model invocation to ingest an entire PLC project together with its accompanying specification. These capabilities pave the way for code generation that encompasses both SFC and ST.

In this paper, we introduce \textsc{\method}, a framework to convert natural language descriptions into executable SFCs. Our approach utilizes a reduced representation and incorporates sub graph masked fine tuning to enhance model training. Additionally, we apply structured generation techniques to ensure compliance with the target grammar. This methodology effectively addresses the challenges associated with generating SFCs from their graphical nature.

\noindent
We summarize our contributions as follows:

\begin{itemize}
    \item To the best of our knowledge, we introduce the first framework for generating machine-executable Sequential Function Charts (SFCs) from natural language descriptions of industrial control workflows.
    \item We present a fine-tuning scheme for generating machine-executable SFCs via LLMs 
    \item We present an extensive analysis of the generation performance across notable open-source and proprietary LLMs.
\end{itemize}

\section{Related Works}

Recent advances in large language models (LLMs) for code generation, exemplified by systems like Qwen models \cite{qwen2.5-coder, qwen3}, and proprietary models like GPT \cite{openai_chatgpt_2025} and Gemini \cite{google_gemini_2025} have increased attention on using LLMs for Programmable Logic Controller (PLC) code. Under the IEC 61131-3 standard, PLC programming can be done in Ladder Diagrams (LD), Function Block Diagram (FBD), Structured Text (ST), Instruction List (IL), and Sequential Function Charts (SFC). Previous work has focused largely on ST generation, thanks to its similarity to high-level languages such as C and C++. 

Among recent efforts, Fakih et al. \cite{LLM4PLC} created a generation pipeline which leverages grammar checkers and model-checking tools such as NuXmv \cite{nuxmv} which feed back information to the LLM regarding the errors in the generated code, and allowing the LLM to correct it's mistakes until it successfully generates syntactically and functionally correct ST code, setting a strong baseline. Another line of research looks at agent-based approaches for automation. In \cite{agent4plc}, different LLM-driven modules, each dedicated to a particular aspect of PLC programming, operate in a loop to iteratively refine the code, addressing syntax and verification issues as they arise.

However, despite these steps forward, comparatively little research has tackled the generation of graphical PLC languages (e.g., SFC). Such work remains limited, partly due to complexities in representing and verifying non-textual elements (e.g., states and transitions) using current generative methods. To our knowledge, the sole prior work is \cite{ascii-sfcs}, which focused on producing ASCII sketches of SFCs rather than machine-executable charts.
As a result, while ST-focused generation pipelines have made significant strides, open questions remain on how to extend these techniques to graphical languages and on how to ensure that the generated diagrams can be validated as reliably as text-based code.

\section{Background}
\begin{figure*}[ht]
    \centering
    \includegraphics[width=1\linewidth]{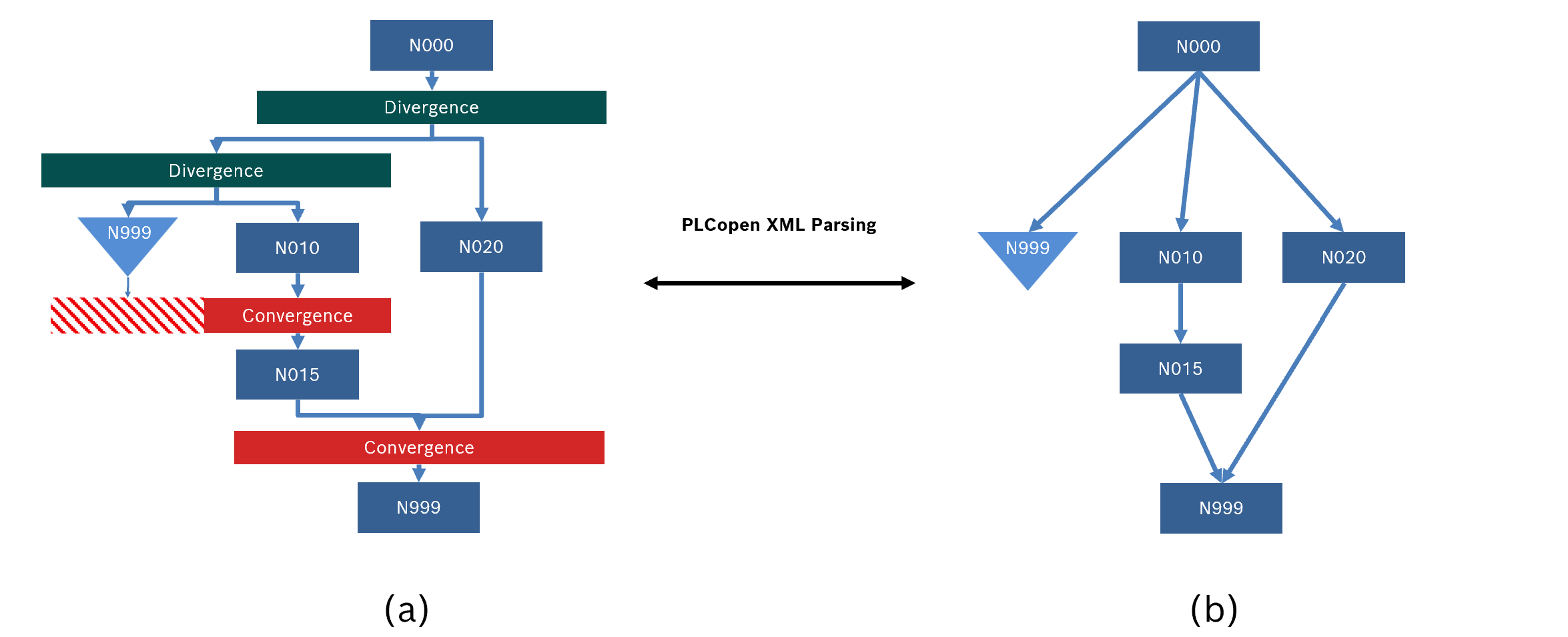}
    \caption{Conversion between PLCopen XML representations and our reduced internal graph structure. (a) An example SFC as encoded in PLCopen XML, including convergence and divergence nodes and vendor-specific metadata. (b) The reduced representation omits redundant metadata and infers structural roles (e.g., branching) from transition semantics.}
    \label{fig:parsing}
    \Description[Parsing between PLCopen and our representation]{
    Conversion between PLCopen XML representations and our reduced internal graph structure. (a) An example SFC as encoded in PLCopen XML, including convergence and divergence nodes and vendor-specific metadata. (b) The reduced representation omits redundant metadata and infers structural roles (e.g., branching) from transition semantics.
    }
\end{figure*}

\subsection{Sequential Function Chart}

Sequential Function Charts are a graphical programming language defined by the IEC 61131-3 standard for programming logic controllers (PLC). Unlike text-based languages, SFCs utilize a flowchart-like visual format to model control processes that operate through discrete steps or states. This makes them particularly effective in applications where processes follow a logical sequence, such as in industrial automation systems or time-based operations. SFCs are also effective for debugging and diagnostics purposes since their visual character is easy to understand and follow along a complex process. SFCs improve system clarity by decomposing complex control logic into modular elements: \textit{steps} (representing states), \textit{transitions} (defining conditions to move between steps), and \textit{actions} (executable instructions in a valid programming language). \textit{Steps} define the current status or active phase in a process, for every SFC there exists exactly one initial step. \textit{Transitions} define the control flow between steps and are labeled with associated transition conditions, commonly referred to as \emph{guards}. Several types of transitions exist, each serving a distinct control function within the chart:
\begin{itemize}
  \item \textbf{Simple transitions}, which connect one step to another directly. Regular transitions always lead downward in the workflow, while \textit{Jump steps} are used if we want to return to a step that precedes the current step. 
  \item \textbf{Alternative branching}, representing a mutually exclusive choice between multiple transitions originating from a common step. 
  \item \textbf{Parallel branching}, which denotes a parallel divergence from a single step into multiple concurrently active branches after which converge into a single step which synchronizes multiple parallel branches back into a single step.
\end{itemize}

\textit{Actions} specify the operations or control signals that are executed when a step is active. Actions can be written in any of the 5 languages defined in the IEC 63113-3 standard but are often implemented in ST and one or more actions may be associated to each step.


    
    
    


\subsubsection{PLCopen XML format}
Although SFCs are graphical in nature, they can also be represented textually through the vendor-neutral PLCopen XML standard.

PLCopen XML is a standardized XML format defined by the \emph{PLCopen} organization \cite{plcopen} to represent automation programs, particularly those written in IEC 61131-3 languages. Using the PLCopen XML standard, SFC structures are encoded as XML elements. The format enables \emph{interoperability} between PLC programming tools by allowing import and export of logic across different vendors. Its main goal is to provide a \emph{vendor neutral, machine-readable} format that supports toolchains, simulation, code generation, and formal verification.
This representation complies with widely used PLC programming editors like TwinCAT \cite{twincat3} and CODESYS \cite{codesys}

\subsubsection{Safe Sequential Function Charts (Safe SFCs)}
\label{subsubsec:safe_sfc}
The IEC 61131-3 syntax guarantees well-formedness but not run-time safety. While permissible by the IEC 61131-3, certain transitions break synchronization or are logically impossible. For example, a transition out of a parallel branch without proper synchronization or jumps between parallel branches. To address logical fallacies of this kind, a \emph{Safe SFC} \cite{safe-sfc} is defined as one that eliminates jumps that cross or exit parallel branches without synchronization and avoids convergences whose source steps can never be marked together; equivalently, no reachable state carries more than one token per branch and every convergence remains attainable. In practice, tools first sweep the graph to spot mismatched divergence, convergence pairs and illegal jumps, then hand a guard-free model to a symbolic checker such as nuXmv, which proves token-overflow freedom and convergence reachability.

\subsection{Large Language Models}

Large Language Models (LLMs) are neural networks trained to model natural language by, most commonly, predicting the next token in a sequence. Most contemporary LLMs are built on the Transformer architecture \cite{attention-is-all-you-need}, which has consistently demonstrated strong performance and efficiency across a wide range of tasks, in particular language.

\subsubsection{Pre-training and Fine-tuning}

Auto-regressive LLMs predict the next token $x_{t+1}$ given a prior token sequence $x_{\leq t} \triangleq (x_1, \ldots , x_t)$ by estimating the probability distribution of the next token, such that:
\begin{equation}
    x_{t+1} \sim P(x_{t+1} | x_{\leq t};\Theta) = f_\Theta(x_{\leq t}),
\end{equation}
where the function $f_{\Theta}(\cdot)$ describes the LLM with parameters $\Theta$. LLMs are typically pre-trained on large corpora by minimizing the negative log likelihood $\mathcal{L(\Theta)}$ of token sequences $x \triangleq (x_1, \ldots, x_t)$ sampled from a training set distribution $D$:
\begin{equation}
    \mathcal{L}(\Theta) = \mathbb{E}_{x \sim D} \left[ - \sum_{t=1}^{T-1} \log f_{\Theta}(x_{\leq t}) \right],  
\end{equation}
then, the LLM's optimal parameters $\Theta^*$ are given as:
\begin{equation}
    \Theta^* = \arg\max_{\Theta} \mathcal{L}(\Theta)
\end{equation}

This auto-regressive training objective equips LLMs with general language capabilities. Fine-tuning adapts pretrained LLMs to specific tasks by updating model weights. However, full fine-tuning is costly for large models. To address this, parameter-efficient techniques such as LoRA have been developed. 

\subsubsection{LoRA Fine-tuning}
LoRA (Low-Rank Adaptation) \cite{lora} introduces a parameter efficient fine-tuning method by keeping the original pretrained weights frozen and injecting trainable low-rank matrices into the pretrained weights.
For a weight matrix \(\theta_0 \in \mathbb{R}^{d \times k}\ \subseteq \Theta \), LoRA defines:
\begin{equation}
    \theta_{L} = \theta_0 + \Delta \theta = \theta_0 + B \cdot A,
\end{equation}
with trainable matrices \(B \in \mathbb{R}^{d \times r}, A \in \mathbb{R}^{r \times k}\), and rank \(r \ll \min(d, k)\). This dramatically reduces the number of trainable parameters while preserving the models original weights, making LoRA highly modular. LoRa fine-tuning in code LLMs is often used with instruction-tuned datasets \cite{instruction-tuning} and Fill-in-the-Middle (FIM) objectives \cite{fill-in-the-middle}. 

FIM is particularly effective in code infilling tasks, where the model predicts missing code spans between given contexts. Importantly, FIM training has been shown to retain left-to-right capabilities, a phenomenon known as the "FIM-for-free" property.

\subsubsection{Retrieval-Augmented Generation (RAG)}
While fine-tuning is a very common method to update an LLMs knowledge regarding a certain task, an issue that still exists is that a standalone LLM's knowledge is frozen once training ends or fine-tuning ends, making it prone to outdated facts and hallucinations and sometimes the fine-tuning even damages original capabilities instilled in the LLM.   
RAG \cite{rag} tackles this by using an external text index that the model can query at run-time.

\begin{enumerate}
    \item \textbf{Retrieve.} A dense search module ranks and returns a handful of passages sliced from webpages, manuals, or papers that best match the user's prompt.  
    \item \textbf{Generate with context.} The model then composes its answer while attending to those passages.
\end{enumerate}

 Updating the retrieval index injects fresh knowledge on demand without expensive retraining, while grounding the answer in retrieved passages sharply reduces hallucinations and increase performance in new tasks the original LLM was not trained on.

\subsubsection{Structured Generation}
Structured generation constrains LLM outputs to follow specific formats, such as XML, JSON or code. While prompting and fine-tuning help guide structure, they do not guarantee structural correctness.
Structured generation methods, such as \cite{outlines, xgrammar}, enforce structure during decoding by using finite-state machines (FSMs) or parsers built from regular expressions or context-free grammars (CFGs). At each generation step, it masks invalid tokens by consulting the FSM or parser state, ensuring that only tokens leading to a valid continuation can be sampled. This guided sampling continues until a valid structured output is complete or context runs out. This approach is particularly relevant to tasks like SFC generation, where the output must adhere to a strict and hierarchical schema.

\section{LLM4SFC Methodology}
    \begin{figure*}
        \centering
        \includegraphics[width=1\linewidth]{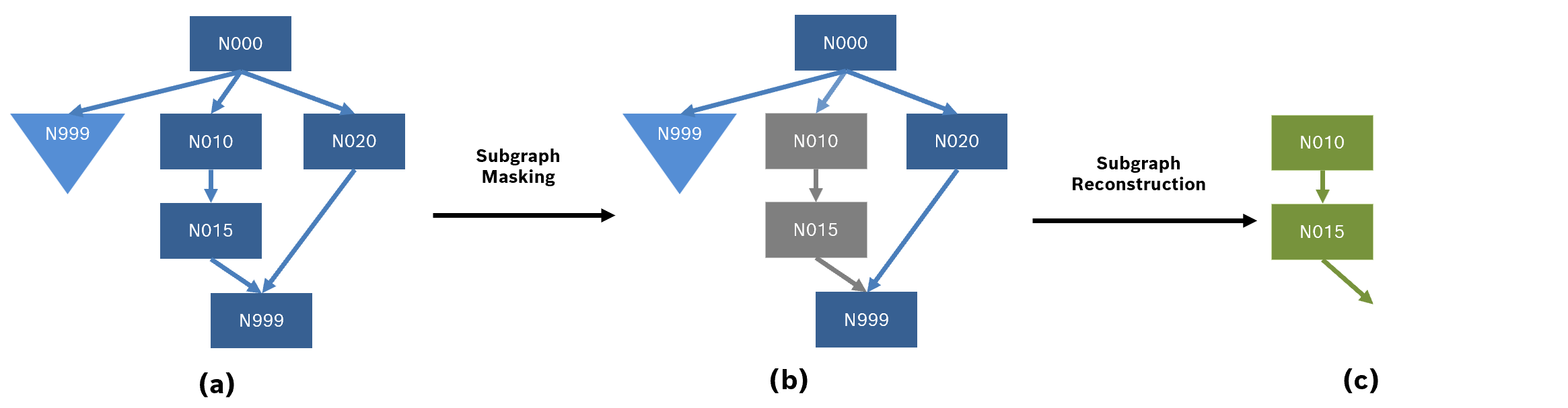}
        \caption{Illustration of our proposed Subgraph Masking fine-tuning strategy. (a) The original reduced SFC graph. (b) A subgraph consisting of structurally related steps and transitions is randomly masked during training. (c) The model is trained to reconstruct the missing subgraph components, promoting a deeper structural understanding of SFC topology. This technique combines graph-masked language modeling and FIM objectives to enhance the model's ability to generate coherent and complete control logic segments.}
        \label{fig:subgraphmasking}
        \Description{An illustration of our proposed Subgraph Masking fine-tuning strategy.}
    \end{figure*}

In the following section, we present \method, an end-to-end workflow for generating Sequential Function Charts from natural language specifications. 
Section \ref{sec:rep_schema} describes the preprocessing pipeline for existing SFCs, in which graphical SFCs are converted into a \emph{reduced textual representation}. This representation can be deterministically recompiled into valid PLCopen standard for downstream use. Section \ref{subsec:dataset-method} outlines the construction of our dataset, including the extraction, augmentation, and generation of the relevant components for each SFC. Section \ref{subsec:fine-tuning} details the fine-tuning strategies used to adapt open-source LLMs to the SFC generation task, while Section \ref{subsec:rag} describes the retrieval-augmented generation pipeline used to condition the model on both the user prompt and relevant retrieved examples. The final output of this workflow is a reduced SFC generated by the LLM, which is then compiled to PLCopen XML and imported into the IDE for simulation or deployment on a physical station (see Fig.\ref{fig:LLM4SFC-pipeline}).

\subsection{SFC Reduced Representation}\label{sec:rep_schema}

SFCs represented via the PLCopen format describes all steps, transitions, convergences, divergences, and the ST code contained within the text. The PLCopen format is highly verbose and contains much redundant metadata, making it unfeasible for an LLM to accurately generate SFCs as it quickly reaches output token limits and requires many decisions from the LLM to be consistently correct. Additionally, previous research \cite{letmespeakfreely} indicates that strict formatting requirements imposed by the cumbersome format can affect performance in structured generation tasks.

We introduce the reduced representation (see Fig \ref{fig:parsing}) which describes the SFC in a format that contains only necessary topological information needed to reconstruct the entire SFC. The reduced SFC is structured as a list of step nodes, containing:
\begin{itemize}
\item Step name.
\item Step 'Action', i.e., its contained ST code.
\item Children step nodes with the respective transition condition, and a flag denoting whether the child is accessed through a jump step.
\item Step comments, which do not affect functionality but assist in readability.
\end{itemize}
In addition, an SFC also contains variable declarations for its entirety, which are divided into input, output, and local variables. Each variable is described via its name, type, and default value. Reducing the textual representation of an SFC makes it more suitable for language model training and inference while also reducing the cost of generation by reducing the input token count thus enabling the generation of larger SFCs.

\subsection{Parsing Between PLCopen and the Reduced Representation} \label{sec:parsing}
Parsing to and from the reduced representation is performed programmatically. We use the \emph{xsdata}\footnote{\url{https://xsdata.readthedocs.io/en/latest/}} Python package for XML parsing and the \emph{plcopen}\footnote{\url{https://pypi.org/project/plcopen/\#description}} Python package as a data class template for loading from and saving to the PLCopen format.

\paragraph{Reduced Representation to PLCopen}
The PLCopen format contains many redundant metadata fields, which are regenerated when the file is loaded in vendor specific IDEs such as TwinCAT. As such, we use minimal nodes from an example project as templates, parsing them into a Python data class. Our representation does not include explicit information about convergence and divergence nodes; instead, we derive their locations, parents, and children within the SFC graph from the transition conditions (parallel branching for identical conditions; otherwise, alternative branching). Eventually, we construct a graph in which each step node (except the initial step) has one parent and one child, and branching is handled via convergence and divergence nodes (see Fig. \ref{fig:parsing}). Finally, we pass the node information to the \textit{plcopen} data class and generate the PLCopen file. The missing metadata is regenerated when the PLCopen file is imported into the intended IDE.

\paragraph{PLCopen to Reduced Representation}
We load the PLCopen file directly into the Python data class provided by \emph{plcopen}; the PLCopen format specifies each node's parent and links each step to its action via an ID field. This allows us to construct a step graph with the appropriate transitions and to associate ST code blocks with their respective steps. We discard the PLCopen metadata, leaving only the information required to reconstruct a functionally identical PLCopen file.

\subsection{Dataset Construction Methodology}\label{subsec:dataset-method}
We constructed a dataset of Sequential Function Charts by collecting code from manufacturing automation projects. Each source SFC was available in PLCopen format and contained multiple steps with associated Structured Text (ST). We parsed the PLCopen SFCs into the reduced SFC representation introduced in Section \ref{sec:rep_schema}.

Textual summaries were generated automatically via GPT-4o, then augmented into  instruction-style prompts form to mimic realistic engineering queries. We embedded the resulting summaries using OpenAI's \texttt{text-embedding-ada-002}\footnote{\url{https://platform.openai.com/docs/models/text-embedding-ada-002}} model so that semantically similar SFCs can be efficiently retrieved at inference time.

Each dataset item therefore contains: (i) raw PLCopen input, (ii) reduced SFC structure, (iii) natural-language description of the SFC (iv) embedding vector.

\subsection{Fine-tuning} \label{subsec:fine-tuning}
    Our fine-tuning methodology is guided by three core objectives:
\begin{enumerate}
    \item To accurately predict the structural topology of SFCs, encompassing the correct configuration of steps, transitions, and actions.
    \item To generate corresponding Structured Text (ST) code that is both syntactically valid and semantically aligned with the control flow, ensuring correct execution semantics.
    \item Both the SFC graph and the contained ST must follow a set of conventions for structure, naming, and comments.
\end{enumerate}
    To address these goals, we incorporate two techniques: \hyperref[par:ntp]{\emph{Next Token Prediction}} and \hyperref[par:subgraphmasking]{\emph{SFC Subgraph Masking}}.

\begin{figure*}[t]
    \centering
    \includegraphics[width=.75\linewidth]{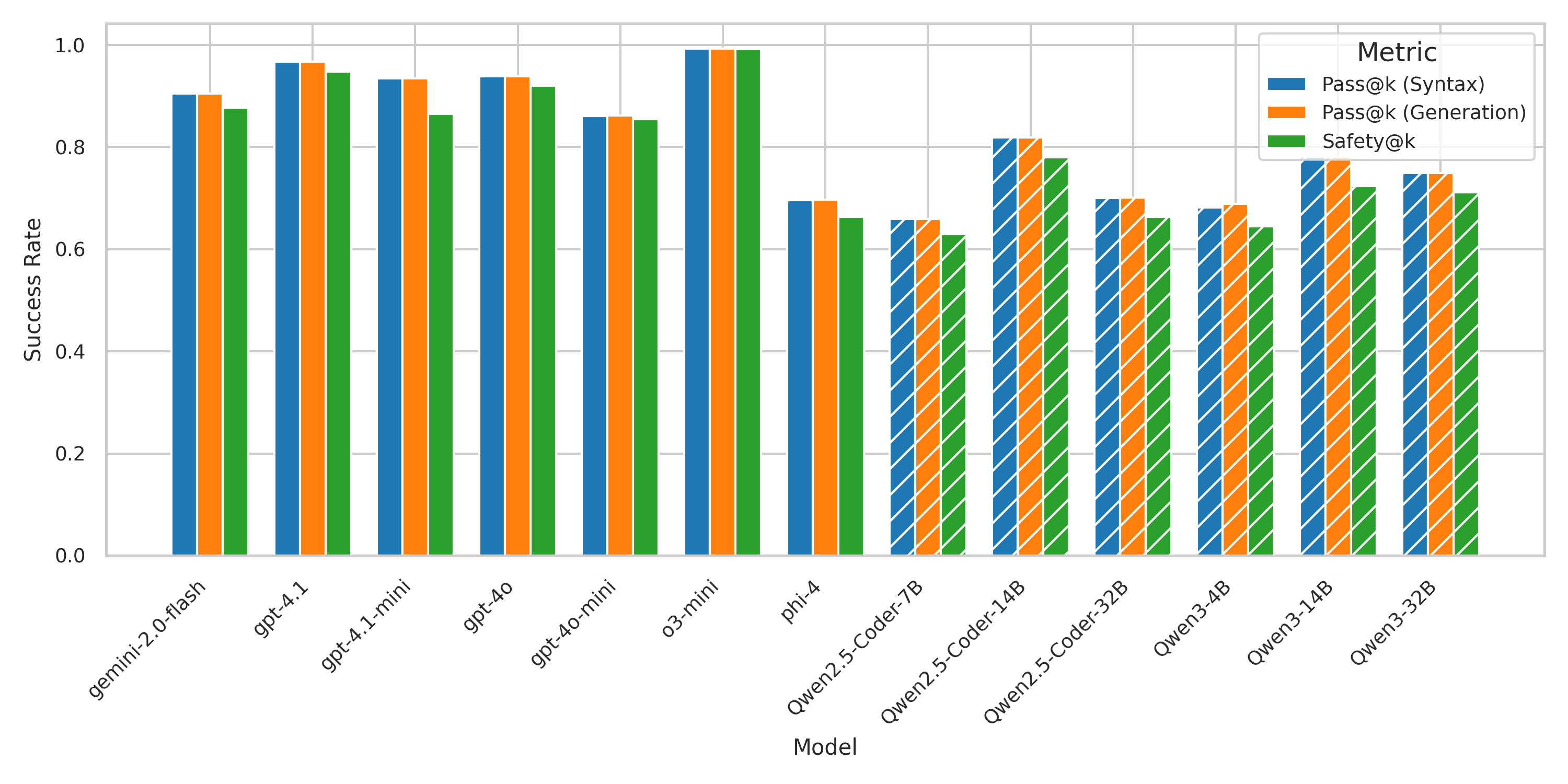}
    \caption{Performance comparison of various models on Pass@k and Safety@k metrics. For each model, we show the results from its best-performing configuration based on the Pass@k (Generation) score. Proprietary models (solid bars) use $k=10$, while open-source models (hatched bars) use $k=5$.
     The plot highlights a general trend of proprietary models outperforming open-source models across these metrics.  Further error analysis of these performance differences is discussed in the section \ref{subsec:error-analysis}.}
    \label{fig:pass_safety}
    \Description{Performance comparison of various models on Pass@k and Safety@k metrics. For each model, we show the results from its best-performing configuration based on the Pass@k (Generation) score. Proprietary models (solid bars) use $k=10$, while open-source models (hatched bars) use $k=5$.}
\end{figure*}

\paragraph{Next Token Prediction} \label{par:ntp}
    The \textbf{next-token prediction (NTP)} follows the standard autoregressive language model formulation: given a prefix of tokens $\mathbf{x}_{<t} = (x_{1},\ldots,x_{t-1})$, the model is trained to maximize the likelihood of the next token $x_{t}$.  
    To apply this objective to our dual-modality data (SFC topology \emph{and} Structured Text code) we treat the SFC as a single token stream and then fine-tune the base model by minimizing the cross-entropy loss across the entire sequence.

\paragraph{SFC Subgraph Masking} \label{par:subgraphmasking}
We adopt a hybrid approach that integrates graph-masked language modeling techniques \cite{maskedlanguagemodeling} with FIM fine-tuning \cite{fill-in-the-middle}. During fine-tuning, we randomly mask subgraphs clusters of steps, transitions, and actions (see Fig. \ref{fig:subgraphmasking}) and train the model to reconstruct them, promoting structural understanding. The model is then trained to reconstruct the masked components.

\subsection{Few-shot Prompting} \label{subsec:rag}
As detailed in section \ref{subsec:dataset-method}, each SFC is summarized into a natural language description. For every description we create a sentence embedding and index it. At inference time, the incoming SFC prompt is embedded with the same model, the $k$ closest vectors in the index are retrieved, and the associated SFC examples are inserted into the model's context as few-shot demonstrations before generating the new chart, where $k$ is a hyperparameter chosen in advance.

\section{Experimental Setup}\label{sec:experimental_setup}

To assess the performance of our proposed framework for SFC generation, we conduct a series of targeted experiments designed to address the following research questions (RQs):

\begin{itemize}
    
    \item RQ1: To what extent can {\method} generate SFC code that is machine-executable, syntactically correct and safe?
    \item RQ2: What is the individual contribution of each component of the framework to the quality of SFC generation?
    \item RQ3: To what extent do the generated SFCs reflect the original intent expressed in the input prompt?  
\end{itemize}

\subsection{Dataset Details}\label{subsec:dataset}
The final dataset comprises 2,390 SFCs drawn from the 34 industrial projects, spanning a range of real-world manufacturing control tasks. For each SFC we store:
\begin{itemize}
  \item Reduced SFC representation (Section \ref{sec:parsing}).
  \item Natural-language summary generated from that representation which is augmented into a instruction-style prompt.
  \item Embedding vector derived from the summary.
\end{itemize}

We split the corpus into training/validation/test partitions using an 80\%/10\%/10\% ratio. The validation set is used for early checkpoint selection during fine-tuning; all reported metrics use the held-out test set.

\subsection{LLM Experimental Setup}
We evaluated \method's performance for two groups of core LLMs:

\textit{Proprietary LLMs}: Per our dataset usage policies, we have access to a selection of proprietary LLMs through an intermediary provider. We evaluate the following models:

\begin{itemize}
    \item \textbf{OpenAI:} \textit{GPT-4o} series, \textit{GPT-4.1}, and \textit{o3-mini}.
    \item \textbf{Google:} \textit{Gemini 2.0 Flash}.
\end{itemize}
The intermediary provider does not offer the option to fine-tune these models at the time of writing. Structured responses are queried via the OpenAI Python interface.

\textit{Open-source LLMs}: We fine-tuned a selection of state-of-the-art open-source LLMs via the process described in Sec.\ref{subsec:fine-tuning} and a combination of the two, i.e. performing NTP and Subgraph masking sequentially. We decided to focus on LLMs which were trained with code generation in mind and also supported Fill-in-the-Middle \cite{fill-in-the-middle} training.
We fine-tuned language models at three different parameter scales:
\begin{itemize}
    \item \textbf{4B models:} \textit{Qwen3-4B} \cite{qwen3}.
    \item \textbf{7B models:} \textit{Qwen2.5-Coder-7B} \cite{qwen2.5-coder}.
    \item \textbf{14B models:} \textit{Microsoft Phi-4} \cite{phi4}, \textit{Qwen3-14B} \cite{qwen3}, and \textit{Qwen2.5-Coder-14B} \cite{qwen2.5-coder}.
    \item \textbf{32B models:} \textit{Qwen3-32B} \cite{qwen3} and \textit{Qwen2.5-Coder-32B} \cite{qwen2.5-coder}.
\end{itemize}
 For each fine-tuned model variant, we conducted two types of generation tasks: (1) free generation, in which the model predicts tokens auto-regressively without constraints, and (2) structured generation, guided by finite state machine (FSM)-based logit manipulation. Finally, we perform separate experiments to compare zero-shot and few-shot generation. In the zero-shot setting, the model is prompted to generate an SFC given only a specific scenario. In the few-shot setting, we first retrieve three semantically similar SFCs, based on the cosine similarity of their textual descriptions and present their reduced representations to guide the generation.
Training and inference were conducted on a compute node equipped with an NVIDIA H200 GPU. We utilized the Unsloth framework \cite{unsloth} for fine-tuning and vLLM \cite{vllm} for efficient inference. In line with prior work, we adopted previously validated hyperparameters: all LoRA adapters were configured with a rank of 128 and scaling factor $\alpha=128$. Optimization was performed using the Adam optimizer with a batch size of 32 and a learning rate of $2 \times 10^{-5}$. For NTP fine-tuning, we trained the model for 5 epochs over the training set. For the subgraph masking objective, we generated three masked examples per SFC and trained the model for 3 epochs on the resulting dataset. 

\begin{figure*}[ht]
    \centering
    \includegraphics[width=.8\linewidth]{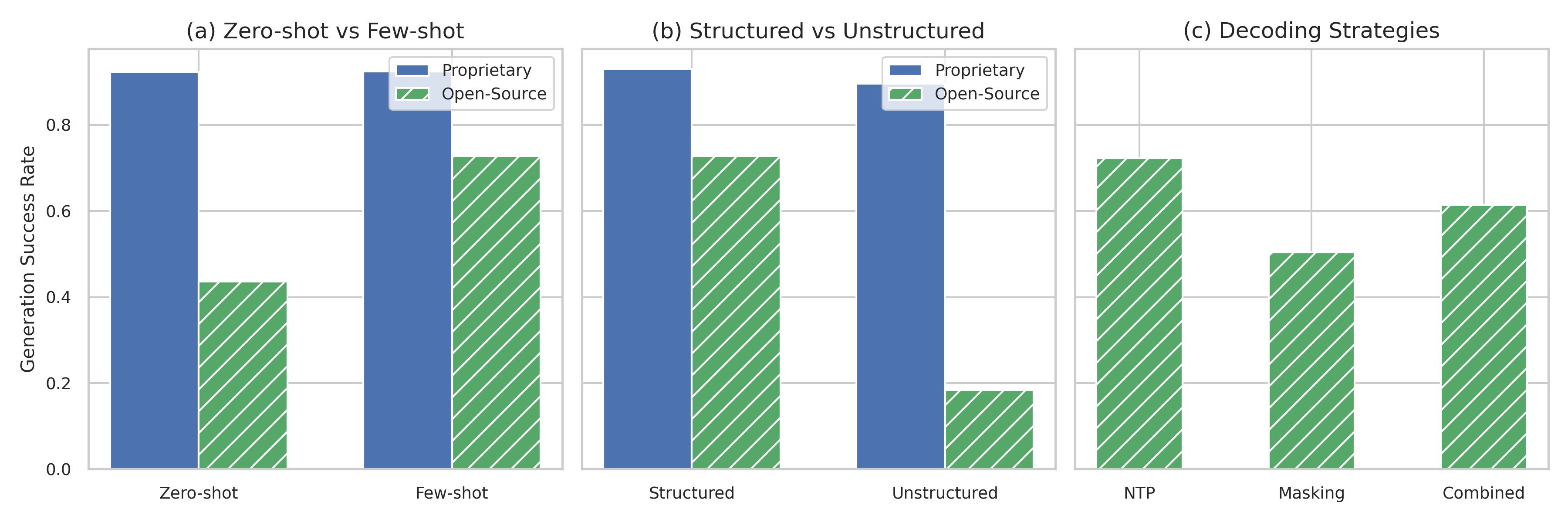}
    \caption{Comparison of generation success rates across different experimental setups. (a) Proprietary models outperform open-source models in both zero-shot and few-shot scenarios. (b) Structured generation improve generation success for both proprietary and open-source models compared to unstructured prompts. (c) Among decoding strategies tested for open-source models, NTP offers only an edge over the masking and combined approaches.}
    \label{fig:ablation}
    \Description{Comparison of generation success rates across different experimental setups, zero-shot vs few-shot, structured vs unstructured decoding, NTP vs Subgraphmasking vs both}
\end{figure*}

\subsection{Metrics}
To automatically evaluate Sequential Function Charts (SFCs) we adopt an evaluation protocol that examines each SFC from three important perspectives. We choose to report the following metrics as we believe that they are of most interest to PLC programmers intending to future code assistance tools built on our framework.  
    
    \subsubsection{Pass Rate@k - Structural and syntactic validity}
    First, the intermediate representation's schema is validated before being translated to the PLCopen XML schema. Every Structured Text (ST) fragment associated with a step or action is extracted, and parsed with the ST parser supplied by \textsc{PLCreX} \cite{plcrex}, an open-source Python package for IEC 61131-3 PLCs, enabling analysis and validation of Structured Text code. It supports ST validation via integrated parsing, IEC-compliance checking. A generation is marked as \emph{passed} if it passes all syntactic and structural validation checks. We report the percentage of valid generations within the top-5 completions.

    \subsubsection{Safety Rate@k - Safety guarantees}
    Control-flow safety is verified against the criteria for \emph{Safe SFCs} put forward by \cite{safe-sfc}. Concretely, we check (i) reachability of all terminal steps from the initial step, (ii) correctness of step-transition connectivity, (iii) validity of jump destinations, and (iv) absence of illegal activations in parallel regions. For each structurally valid SFC we model the transition system via the NuXmv software \cite{nuxmv} and verify the safety criteria. An SFC is considered safe if it has passed all safety checks listed in the \ref{subsubsec:safe_sfc}. We report the percentage of safe generations within the top-$k$ completions.

    \subsubsection{MATCH Score - Semantic alignment}
    To quantify how faithfully a generated chart realizes the intent expressed in natural language, we compute a neural similarity metric via the MATCH score \cite{match2025}. 
    MATCH scores have been shown to correlate strongly with human judgments of functional adequacy, therefore we use MATCH score to showcase that the generated chart appropriately satisfies the user's request. A semantic alignment metric measuring the similarity between the generated chart and the input prompt, averaged over the test set. We trained and validated the MATCH architecture on the train and validation splits of our dataset.

\section{Results}\label{sec:results}

In this section, we present the empirical findings of our experiments, structured according to the three research questions introduced in Section \ref{sec:experimental_setup}. Results are reported for both proprietary and fine-tuned open-source models, across structured and unstructured generation modes, and under both zero-shot and few-shot prompting conditions.

\subsection{RQ1: Executability, Syntax, and Safety}
We conducted an evaluation across a range of proprietary and open-source coding models. We report  the performance of each model as it was evaluated under few-shot prompting with structured generation enabled, as these conditions yielded the strongest results during preliminary benchmarking \ref{tab:proprietary-results}. The few-shot setting provided the necessary context for the model to adhere to the SFC format and aligned the generation with the conventions of SFC programming in our dataset, while structured decoding helped enforce format specific constraints during generation.

As shown in Figure \ref{fig:pass_safety} and Tables \ref{tab:open-source-results} and \ref{tab:proprietary-results} , proprietary models (e.g.,GPT-o3-mini, GPT-4o-mini, Gemini 2.0 Flash) substantially outperformed their open-source counterparts across all metrics. GPT-4o-mini and GPT-4.1 achieved strong performance with over 95\% on both syntax and generation metrics, and Safety@k scores exceeding 80\%. While proprietary models achieve near perfect performance, open-source models like Phi-4 and Qwen2.5-Coder-7B exhibited much lower reliability, particularly in safety validation, often scoring below 70\%. Nonetheless, we anticipate that these results will continue to improve as fine-tuning becomes more accessible and open-source code generation models rapidly advance, paving the way for broader adoption and increased reliability in SFC generation.

\begin{table}[ht!]
  \centering
  \resizebox{\linewidth}{!}{
    \begin{tabular}{lccccc}
      \toprule
      \textbf{Model} & \makecell{\textbf{Fine Tuning} \\ \textbf{Type}} &
      \makecell{\textbf{Pass@10} \\ \textbf{(Gen)}} &
      \makecell{\textbf{Pass@10} \\ \textbf{(Syntax)}} &
      \textbf{Safe@10} \\
      \toprule
        phi-4 & NTP & 69.70\% & 69.58\% & 66.36\% \\
        Qwen2.5-Coder-7B & NTP & 65.90\% & 65.90\% & 62.97\% \\
        Qwen2.5-Coder-14B & NTP  & \textbf{81.88}\% & \textbf{81.88}\% & \textbf{77.99}\% \\
        Qwen2.5-Coder-32B & NTP + Masking & 70.08\% & 70.04\% & 66.32\% \\
        Qwen3-4B & NTP & 68.87\% & 68.16\% & 64.48\% \\
        Qwen3-14B & NTP & 78.12\% & 78.07\% & 72.34\% \\
        Qwen3-32B & NTP & 74.93\% & 74.89\% & 71.13\% \\
      \bottomrule
    \end{tabular}%
  }
  \caption{Pass@5 - generation success rate, ST syntax rate, and safety rate for the best performing configuration for each fine-tuned open-source model, Higher is better. Across all models, the combination of 3-shot prompting and structured generation consistently outperformed alternative configurations.
  }
  \label{tab:open-source-results}
\end{table}

These results suggest that while proprietary models are currently capable of generating executable and syntactically valid SFCs at high reliability, most open-source models require further fine-tuning or architectural adaptations to MATCH this level of robustness. Nonetheless, some structured open-source models (e.g., Qwen2.5-Coder-14B) show promise, indicating that further refinement, such as prompt engineering or hyperparameter tuning, could narrow the performance gap.
Tables \ref{tab:open-source-results} and \ref{tab:proprietary-results} present the Pass@k scores for generation success, syntax conformity, and safety across both open-source and proprietary models. For open-source models, we report results with $k=5$, whereas for proprietary models, we use $k=10$. Although our original intention was to use $k=10$ on both, generating 10 samples per prompt with open-source models proved computationally infeasible due to hardware constraints. In contrast, the primary limitation for proprietary models was API rate limits, which were comparatively easier to manage.

\begin{table}
\centering
\begin{tabular}{lccc}
\toprule
 \textbf{Model} & \makecell{\textbf{Pass@10} \\ \textbf{(Gen)}} & \makecell{\textbf{Pass@10} \\ \textbf{(Syntax)}} & \textbf{Safe@10} \\
\toprule
gemini-2.0-flash (S,3) & 94.96\% & 94.92\% & 85.59\% \\
gpt-4.1 (S,3) & 96.74\% & 96.74\% & 94.77\% \\
gpt-4.1-mini (S,3) & 93.43\% & 93.43\% & 86.48\% \\
gpt-4o (S,0) & 93.84\% & 93.84\% & 92.00\% \\
gpt-4o-mini (U,3) & 86.15\% & 86.10\% & 85.50\% \\
o3-mini (S,0) & \textbf{99.25}\% & \textbf{99.25}\% & \textbf{99.16}\% \\
\bottomrule
\end{tabular}
\caption{Pass@10 results showing generation success rate, structured text (ST) syntax conformity, and safety for the best-performing configuration of each proprietary model. Structured generation is denoted by \textit{S}, unstructured by \textit{U}, few-shot (3-shot) prompting by \textit{3}, and zero-shot by \textit{0}}
\label{tab:proprietary-results}
\end{table}

\subsection{RQ2: Component Effect Comparison - Fine-Tuning and Generation Strategies}

We conducted an ablation study focusing on three axes of variation: \textbf{prompting strategy} (zero-shot vs. few-shot), \textbf{decoding strategy} (structured vs. unstructured decoding), and \textbf{fine-tuning mechanism} (Next-Token Prediction, subgraph-masking, and NTP + Masking). Each configuration was evaluated in isolation to quantify its effect on the \textit{generation success rate}, the proportion of generated outputs that were valid PLCopen XML.

All experiments were conducted using a fixed temperature of $0.3$, in accordance with prior studies indicating that higher temperatures can negatively impact code generation performance \cite{low-temp}. Aside from temperature, all remaining sampling parameters were configured according to the recommendations provided by the respective LLM provider. Zero-shot and few-shot prompting used the same base task prompt, with the few-shot condition appending 3 example SFCs to guide model behavior. Structured decoding enforced outputs to remain within the expected reduced representation format, while unstructured decoding allowed free-form completions. Fine-tuning comparisons were performed only on open-source models due to us not having fine-tuning access to proprietary models. Results are presented in Figure \ref{fig:ablation}, highlighting three key findings:

\textbf{(1) Few-shot prompting improves open-source performance.} As shown in Figure \ref{fig:ablation} (a), open-source models saw a substantial gain in generation success rate when moving from zero-shot ($\sim$40\%) to few-shot prompting ($\sim$70\%). Proprietary models benefited slightly, although they performed strongly in both conditions ($>$90\%).

\textbf{(2) Structured decoding is essential for open-source models.} Figure \ref{fig:ablation} (b) demonstrates that enforcing structure improved open model success rates from $\sim$15\% to $\sim$70\%. This indicates that unconstrained decoding is inadequate for generating well formed, machine executable SFCs. Proprietary models also degraded from unstructured output, though the effect was less severe.

\textbf{(3) Fine-tuning strategy influences performance, with NTP slightly outperforming other strategies.} Among open-source models, Figure \ref{fig:ablation} (c) compares decoding strategies. NPT marginally outperforms masking and combined strategies, but differences remain modest. This suggests that while decoding style matters, its effect is secondary to prompting and output format.

In summary, proprietary models show robustness across decoding variations, while open-source models benefit markedly from structured generation and few-shot prompting. These insights inform both practical deployment strategies and directions for further improving SFC generation pipelines.

\subsection{RQ3: Semantic Alignment with Prompt Intent}

To assess the semantic alignment between the users prompt and the generated SFC, we employ the \textbf{MATCH score} \cite{match2025}, a neural similarity metric for code evaluation. The MATCH score quantifies how faithfully the functional behavior described in the natural language prompt is reflected in the output SFC. A MATCH score is computed only for cases where an SFC was generated; failures are excluded and syntax or logical validity is not required. 

We follow the dual-encoder architecture proposed in the original MATCH metric which is comprised of a \texttt{bert-base-uncased} model for textual prompts and a \texttt{microsoft/graphcodebert-base} model for SFC code, augmented with a cross-attention embedding enhancement layer. Training was conducted with a learning rate of $3 \times 10^{-5}$ for up to 50 epochs on the train and validation sets of the dataset described in section \ref{subsec:dataset}. 

As shown in Table \ref{tab:best_success_MATCHscore}, all models achieve high MATCH in their best configurations, with open-source models show slightly higher MATCH score than proprietary models, though with lower generation success. We theorize that this is due to the fact that fine-tuning 

This gap may reflect the benefit of fine-tuning on a corpus of structurally consistent SFCs, which sharpens alignment with human annotation conventions. 

\begin{table}
\centering
\begin{tabular}{lrr}
\toprule
Model & Generation Success & MATCH Score \\
\midrule
\multicolumn{3}{l}{\textbf{Proprietary Models}} \\
gemini-2.0-flash & 94.9\% & \textbf{0.921} \\
gpt-4.1 & 96.7\% & 0.901 \\
gpt-4.1-mini & 93.4\% & 0.911 \\
gpt-4o & 93.5\% & 0.893 \\
gpt-4o-mini & 86.1\% & 0.899 \\
o3-mini & \textbf{99.2}\% & 0.914 \\
\midrule
\multicolumn{3}{l}{\textbf{Open-Source Models}} \\
phi-4 & 69.7\% & 0.9203 \\
Qwen2.5-Coder-7B & 65.9\% & 0.9207 \\
Qwen2.5-Coder-14B & \textbf{81.8}\% & 0.9183 \\
Qwen2.5-Coder-32B & 70.1\% & \textbf{0.9208} \\
Qwen3-4B & 68.9\% & 0.9195 \\
Qwen3-14B & 78.1\% & 0.9192 \\
Qwen3-32B & 74.9\% & 0.9201 \\
\bottomrule
\end{tabular}
\caption{Generation success and MATCH scores for each model in its best configuration. MATCH is computed only on cases where an SFC was produced, syntax or logical validity not required. Open-source models slightly outperform proprietary models on MATCH despite lower generation success rates. Best generation success and best MATCH scores are in \textbf{bold}}
\label{tab:best_success_MATCHscore}
\end{table}

\section{Discussion}\label{sec:discussion}
\subsection{Safety Errors Analysis} \label{subsec:error-analysis}
To better understand the nature of generation failures we conducted an error analysis of the generated SFCs. By categorizing each failure into structural, transition, safety errors, we aim to identify which aspects of SFC generation present the greatest challenges for current models. This breakdown enables a more targeted evaluation of where models struggle and provides insight into the underlying challenges of learning representation of graphical structures.

Each generated SFC was evaluated against a sequence of three conditions, ordered by increasing strictness and difficulty of satisfaction: (1) \textit{Exactly one initial step}, although IEC 61131-3 allows multiple initial steps, we enforce a single initial step since in practice it improves safety by having a single, deterministic entry point; (2) \textit{Complete transition targets}, every transition leaving a step must have a defined destination step; and (3) \textit{Adherence to the Safe SFC guidelines} described in Section \ref{subsubsec:safe_sfc}. Any generated SFC that fails one or more of these checks is flagged as unsafe.

As discussed in the Safe SFC literature \cite{safe-sfc}, deciding structural safety properties by model checking can incur combinatorial blow-up: the explored state space may scale exponentially with the number of steps and parallel branches. Consequently, verification can become prohibitively slow for large or ill structured SFCs. We therefore enforce a maximum verification time of 6 hours per SFC. Runs exceeding this limit are treated as verification failures to maintain a conservative safety posture. 

\begin{table}[htbp]
\centering
\small
\begin{tabular}{lrrrrr}
\toprule
Model  & \makecell{Init Step\\ Error}  & \makecell{Transition \\ Error}  & \makecell{Safety \\ Error}  & \makecell{Timeout \\ Error}  \\
\midrule
\multicolumn{5}{l}{\textbf{Proprietary Models}} \\
gemini-2.0-flash  & 15.87\% & 30.16\% & 41.27\% & 12.70\%  \\
gpt-4.1           & 8.51\%  & 25.53\% & 19.15\% & 46.80\%  \\
gpt-4.1-mini      & 6.02\%  & 45.78\% & 6.62\%  & 41.57\%  \\
gpt-4o            & 43.18\% & 50.00\% & 6.82\%  &   --     \\
gpt-4o-mini       &   --    & 35.71\% & 64.28\% &   --     \\
o3-mini           & 50.00\% & 50.00\% &   --    &   --     \\

\midrule
\multicolumn{5}{l}{\textbf{Open-Source Models}} \\
phi-4                 &  2.66\% & 22.66\% & 12.00\% & 65.33\%  \\
Qwen2.5-Coder-7B      &  9.37\% & 46.87\% & 21.87\% & 31.25\%  \\
Qwen2.5-Coder-14B     &  3.33\% & 17.78\% & 13.33\% & 68.89\%  \\
Qwen2.5-Coder-32B     &   --    & 33.71\% & 30.33\% & 35.96\%  \\
Qwen3-4B              &  2.32\% & 62.79\% & 19.76\% & 17.44\%  \\
Qwen3-14B             &  2.24\% &  5.97\% & 28.36\% & 65.67\%  \\
Qwen3-32B             &   --    &  11.11\% & 10.00\% & 78.89\%  \\
\bottomrule
\end{tabular}
\caption{Safety performance for each model at its best configuration. Percentages show each error's share of that model's total; Cells with "--" indicate no data reported for that error type.}
\label{tab:logic_by_model_best_success}
\end{table}

Table \ref{tab:logic_by_model_best_success} highlights several notable insights into the safety related errors observed in the generated SFCs. Proprietary models such as GPT-4o show a disproportionately high rate of initial step and transition errors (43.18\% and 50.00\%, respectively), suggesting these models still struggle significantly with correctly initializing SFC entry points and defining coherent transition paths. Conversely, some open-source models, exemplified by Qwen2.5-Coder-14B, Qwen3-15B, Qwen3-32B and phi-4, predominantly experience verification timeout errors (>65\%), indicating a tendency to produce overly complex and large SFCs. The second most common error type is the transition error, indicating that the model can still improve in topological understanding of the SFCs. 

Previous works in PLC programming have effectively utilized multi-agent frameworks incorporating automated error detection and correction mechanisms. Building on these methodologies, future advancements in SFC generation can leverage similar automated correction strategies, such as SMV based verification, to mitigate these types of errors and improve model robustness.

\subsection{Limitations}

This pipeline presents several limitations that should be acknowledged:

\paragraph{Data Availability.}
A key limitation of this pipeline is the lack of publicly available datasets containing graphical PLC programming languages, which are critical for both fine-tuning and RAG. This data scarcity constrains the framework from being adopted into a widely used tool but rather a method for commercial companies to implement internally. Furthermore, our work focuses on SFCs specific to manufacturing tasks, whereas SFCs are employed across a broad range of domains, each with its own design conventions and semantics. As such, the models trained in this study are unlikely to generalize effectively to SFCs outside this domain. Nonetheless, the underlying framework is extensible and can support further fine-tuning to accommodate previously unseen SFCs. 

\paragraph{Visual Semantics.}
A core advantage of SFCs is their graphical representation, which improves readability and interpretability for engineers. However, LLMs lack visual perception and are unable to assess the visual clarity of generated SFCs. As a result, LLM-generated SFCs may be functionally correct but visually suboptimal. Future work should include evaluation components assessing an SFCs readability and interpretability in the eyes of a PLC engineer.

\section{Conclusions and Future Work}\label{sec:conclusions}
In this work, we introduced {\method}, the first framework to generate machine-executable Sequential Function Charts from natural language descriptions. Our approach successfully addresses the core challenges of SFC generation by combining a reduced textual representation, targeted fine-tuning strategies, and structured decoding. The reduced representation simplifies the complex PLCopen XML format, making it more amenable for LLMs to learn and generate. Fine-tuning with next-token prediction and our novel subgraph masking technique, alongside few-shot prompting, enables models to capture both the high-level topology and the low-level conventions of SFC programming.

Our evaluation on a real-world dataset of industrial SFCs demonstrates that {\method} can reliably produce syntactically valid and safe programs. Our experiments reveal that techniques like structured generation and few-shot prompting are critical for improving open-source model performance, boosting success rates. Furthermore, our semantic alignment analysis using the MATCH score confirms that both proprietary and open-source models consistently generate SFCs that are faithful to the user's original intent.

The primary limitation of this work is the scarcity of publicly available, high-quality datasets for graphical PLC languages, which are essential for both fine-tuning and retrieval-augmented generation. Future work should focus on expanding data collection and curation efforts. Additionally, exploring multi-modal models that can directly process graphical representations of SFCs could offer a promising avenue for bypassing the need for textual conversion. By bridging the gap between natural language and graphical PLC programming, {\method} paves the way for more automated, efficient, and accessible industrial programming.

\bibliographystyle{ACM-Reference-Format}
\bibliography{main_bib}

\appendix

\end{document}